\documentclass[10pt,twocolumn,letterpaper]{article}

\usepackage{wacv}
\usepackage{times}
\usepackage{epsfig}
\usepackage{graphicx}
\usepackage{amsmath}
\usepackage{amssymb}
\usepackage{balance}
\usepackage{booktabs}

\wacvfinalcopy 

\ifwacvfinal
\def\assignedStartPage{9876} 
\fi

\ifwacvfinal
\usepackage[breaklinks=true,bookmarks=false]{hyperref}
\else
\usepackage[pagebackref=true,breaklinks=true,colorlinks,bookmarks=false]{hyperref}
\fi

\ifwacvfinal
\else
\fi

\begin{document}

\title{Advancing Non-Contact Vital Sign Measurement using Synthetic Avatars}

\author{Daniel McDuff, Javier Hernandez\\
Microsoft, Redmond, USA\\
{\tt\small \{damcduff,javierh\}@microsoft.com}
\and
Xin Liu\\
University of Washington, Seattle, USA\\
{\tt\small xliu0@cs.washington.edu}
\and
Erroll Wood, Tadas Baltrusaitis\\
Microsoft, Cambridge, UK\\
{\tt\small \{erwood,tabaltru\}@microsoft.com}
}

\maketitle

\begin{abstract}
Non-contact physiological measurement has the potential to provide low-cost, non-invasive health monitoring. However, machine vision approaches are often limited by the availability and diversity of annotated video datasets resulting in poor generalization to complex real-life conditions. To address these challenges, this work proposes the use of synthetic avatars that display facial blood flow changes and allow for systematic generation of samples under a wide variety of conditions. Our results show that training on both simulated and real video data can lead to performance gains under challenging conditions. We show state-of-the-art performance on three large benchmark datasets and improved robustness to skin type and motion.
\end{abstract}

\begin{figure*}[h]
  \includegraphics[width=1\textwidth]{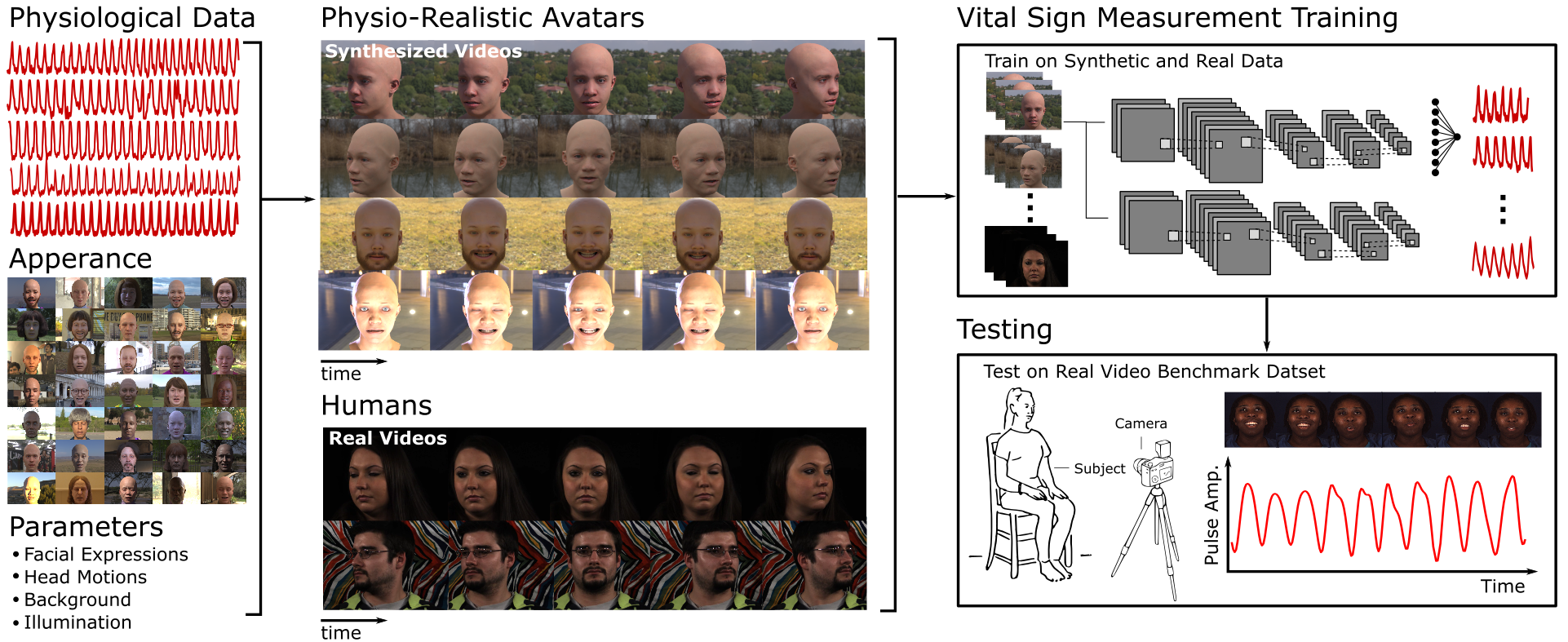}
  \caption{We propose the use of synthetic avatars to improve non-contact physiological measurement via imaging photoplethysmography. Our approach leverages a physically-based model of the subsurface absorption and scattering of light in the skin to display facial blood flow under different conditions: varied motions, backgrounds and appearances.}
  \label{fig:teaser}
\end{figure*}

\section{Introduction}
Photoplethysmography (PPG) is a non-invasive method for measuring peripheral hemodynamics and vital signals such as Blood Volume Pulse (BVP) via light reflected from, or transmitted through, the skin. While traditional PPG sensors are used in contact with the skin, digital imagers can be used offering some unique benefits~\cite{takano2007heart,verkruysse2008remote,poh2010non,blackford2016long}. First, for subjects with delicate skin (e.g.,~infants in a NICU, burn patients, or the elderly) contact sensors can damage their skin, cause discomfort, and/or increase their likelihood of infection. Second, cameras are ubiquitous (available on many tablets, personal computers, and cellphones) offering unobtrusive and pervasive health monitoring~\cite{villarroel2019non}. Third, unlike traditional contact measurement devices (e.g.,~a smartwatch) remote cameras allow for spatial mapping of the pulse signal that can be used to approximate pulse wave velocity and capture spatial patterns in the peripheral hemodynamics~\cite{shao2014noncontact,kamshilin2011photoplethysmographic,kumar2016pulsecam}. 

While there are many benefits of non-contact PPG measurement (a.k.a.,~imaging photoplethysmography (iPPG)~\cite{mcduff2015survey}), this approach is especially vulnerable to different environmental factors posing relevant research challenges. For instance, recent research has focused on making iPPG measurements more robust under dynamic lighting and motion~\cite{wang2017algorithmic,mcduff2018fusing}, and characterizing and combating the effects of video compression~\cite{mcduff2017impact,nowara2019combating,yu2019remote}. 
Historically, iPPG methods often relied on unsupervised methods (e.g.,~ICA or PCA)~\cite{poh2010non,mcduff2014improvements} or hand-crafted demixing algorithms~\cite{CHROMdeHaan,wang2017algorithmic}. Recently, supervised neural models have been proposed providing state-of-the-art performance in the context of heart rate measurement~\cite{chen2018deepphys,yu2019remote,mcduff2018deep,liu2020multi,lee2020meta}. These performance gains are often a direct result of the model scaling well with the volume of training data; however, as with many tasks the volume and diversity of the available data soon become the limiting factor.

Collecting high-quality physiological data presents numerous challenges. First, recruiting and instrumenting participants is often expensive and requires advanced technical expertise which severely limits its potential volume. 
Second, training datasets that have already been collected may not contain the types of motion, illumination changes, or appearances that feature in the application context. Thus, a model trained on these data may be brittle and not generalize well.
Third, the data can reveal the identity of the subjects and/or sensitive health information. For imaging methods this is exacerbated by the fact that most datasets of video recordings include the subjects face in some or all of the frames~\cite{estepp2014recovering,zhang2016multimodal,soleymani2011multimodal}. If we could use synthetic data to train iPPG systems it would, to an extent, side-step all three of these challenges and make for an attractive prospect. Once a graphics pipeline is in place, generation of synthetic data is much more scalable than recording videos. 
In addition, rare events or typically underrepresented populations can be simulated in videos, assuming we have some knowledge of the statistical properties of the events or a set of examples. Furthermore, synthetic datasets would not need to contain faces or physiological signals with the likeness of any specific individual. Finally, parameterized simulations allow us to systematically vary certain variables of interest (e.g.,~velocity of motion or intensity of the illumination within a video) which is both useful to train more robust methods as well as evaluating performance under different conditions~\cite{mcduff2019identifying,vazquez2014virtual}.

We propose to use high-fidelity computer simulations to augment training data that can be used to improve non-contact iPPG measurement (see Fig.~\ref{fig:teaser}). This involves answering several research questions: Can we simulate sufficiently high-fidelity data for training iPPG algorithms? Do model parameters learned on synthetic data generalize to real videos? Can using synthetic data help improve generalizability of the learned model? We hypothesize that this is indeed the case and that data synthesis will play a more important role when creating future non-contact physiological measurement methods. The main contributions of this paper are to: 1)~propose an approach for synthesizing avatars with realistic facial blood flow as synthetic data for training non-contact physiological measurement models, 2)~evaluate a set of models trained on combinations of real and synthetic data on benchmark datasets, 3)~show empirical results that synthetic data can help improve overall performance and offer improvements in cases where data is underrepresented in real-world datasets (e.g.,~task specific motions, or people with darker skin types).

\section{Related Work}
\textbf{Non-Contact Physiological Measurement.}
The BVP can be measured via the light reflected from, or transmitted through, the skin~\cite{allen2007photoplethysmography}. Imaging-PPG is a set of techniques for measuring this signal using non-contact imagers (e.g.,~a~webcam) and ambient light. 
Research has focused on making these algorithms more robust to motion (e.g.,~rigid head motions and speech)~\cite{mcduff2018fusing} and dynamic illumination~\cite{wang2017algorithmic}. Imaging PPG enables the non-contact measurement of several important vital signs and physiological signals including: heart rate~\cite{poh2010non}, respiration~\cite{poh2010advancements,tarassenko2014non}, heart rate variability~\cite{poh2010advancements}, pulse transit time~\cite{shao2014noncontact}, and blood oxygen saturation~\cite{tarassenko2014non}. 
Several datasets have been collected and shared with the research community~\cite{soleymani2011multimodal,zhang2016multimodal,niu2018vipl,bobbia2019unsupervised}. These datasets contain hundreds of videos with ground-truth physiological recordings (either PPG, ECG or both).  However, despite the size and availability of these data there remain limitations. The diversity in skin types, systematic variations of noise signals (e.g.,~motion or lighting changes), and the presence of physiological abnormalities (e.g.,~arrhythmias) are not very high.

\textbf{Training-based on Simulation.}
One of the most notable properties of neural models is how they scale efficiently with the number of training examples. A large amount of engineering and research efforts have been invested in scaling learning infrastructures so that models with vast numbers (millions or billions) of parameters can be trained with time efficiency. However, it is becoming increasingly difficult to collect sufficient volumes of labeled data to exploit this scale, especially for video-based applications.

Using parameterized graphics simulations to augment existing datasets have been extensively explored in different computer vision domains~\cite{shotton2011real,veeravasarapu2015model,veeravasarapu2015simulations,veeravasarapu2016model,vazquez2014virtual,mcduff2019identifying,haralick1992performance} such as training pose recognition~\cite{shotton2011real}, scene segmentation for self-driving cars~\cite{ros2016synthia}, improving object recognition~\cite{peng2015learning}, detecting pedestrians under different conditions~\cite{vazquez2014virtual}, and for performance evaluation of learned models~\cite{haralick1992performance}. AirSim is a graphics-based simulation environment~\cite{shah2018airsim} that has been successfully used in the context of training autonomous drone navigation~\cite{bondi2018airsim} and the systematic evaluation of face detection systems~\cite{mcduff2019identifying}. In the context of physiological sensing; however, synthetic data has been mostly used for evaluation purposes of different algorithms considering other modalities (e.g.,~\cite{Donoso2013, Paalasmaa2014, Charlton2016}). To the best of our knowledge, our work is the first example of using high-fidelity physiological simulations to train iPPG methods. 
Our work is made possible thanks to the ability to render high-fidelity frames/videos with an optical basis for manipulating blood volume in the skin.
Creating realistic blood flow simulations is achieved by modelling the appearance of multiple translucent skin layers~\cite{d2007efficient,jimenez2010real,alkawaz2017oxygenation}. These dynamic appearance models usually capture the subsurface scattering that occurs when light interacts with the outer layers of the skin, and are motivated by in-vivo measurements of melanin and hemoglobin concentrations~\cite{jimenez2010practical}.
We propose that synthetic data can be successfully used for training iPPG systems and leverage these innovations in rendering. We create synthetic data to show how non-contact vital sign measurement can be improved using synthetic data.

\begin{figure*}[h]
  \centering
  \includegraphics[width=1\linewidth]{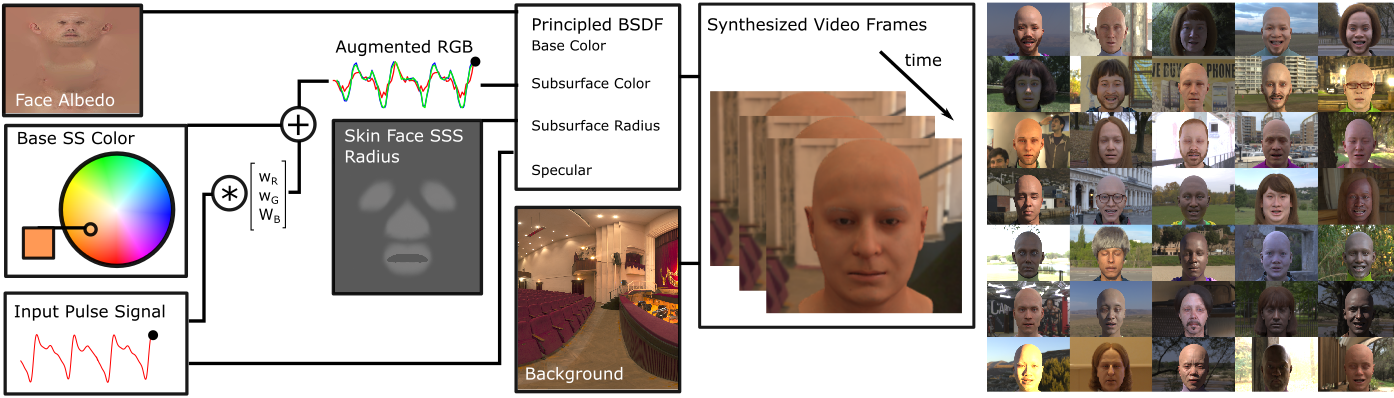}
  \caption{Our approach to synthesizing videos of faces with dynamic blood flow signals. We start with a face albedo, base subsurface color and input pulse signal.  The skin properties are varied temporally based on hemoglobin properties.  The subsurface skin color captures changes in absorption, $\pmb{v}_{abs}(t)$, with variations in hemoglobin. The subsurface scattering, $\pmb{v}_{sub}(t)$, captures how light scattering changes with the volume of blood.}
  \label{fig:approach}
\end{figure*}

\section{Optical Basis for Synthesized Data}

Camera-based vital sign measurement using photoplethysmography involves capturing subtle color changes in skin pixels. Our graphics simulation is inspired by Shafer's dichromatic reflection model (DRM) \cite{wang2017algorithmic}. We start by assuming there is a light source that has a constant spectral composition but varying intensity, the RGB values of the $k$-th skin pixel in an image sequence can then be defined by a time-varying function:
\begin{equation} \label{eq:1}
	\pmb{C}_k(t)=I(t) \cdot (\pmb{v}_s(t)+\pmb{v}_d(t))+\pmb{v}_n(t)
\end{equation}
\begin{equation} \label{eq:1b}
	\pmb{C}_k(t)=I(t) \cdot (\pmb{v}_s(t)+\pmb{v}_{abs}(t)+\pmb{v}_{sub}(t))+\pmb{v}_n(t)
\end{equation}
where $\pmb{C}_k(t)$ denotes a vector of the RGB color channel values; $I(t)$ is the luminance intensity level, which changes with the light source as well as the distance between the light source, skin tissue and camera; $I(t)$ is modulated by two components in the DRM: specular (glossy) reflection $\pmb{v}_s(t)$, mirror-like light reflection from the skin surface, and diffuse reflection $\pmb{v}_d(t)$. 
The diffuse reflection in turn has two parts: the absorption $\pmb{v}_{abs}(t)$ and sub-surface scattering of light in skin-tissues $\pmb{v}_{sub}(t)$; $\pmb{v}_n(t)$ denotes the quantization noise of the camera sensor.
$I(t)$, $\pmb{v}_s(t)$ and $\pmb{v}_d(t)$ can all be decomposed into a stationary and a time-dependent part through a linear transformation \cite{wang2017algorithmic}:

\begin{equation} \label{eq:2}
	\pmb{v}_d(t) = \pmb{u}_d \cdot d_0 + (\pmb{u}_{abs} + \pmb{u}_{sub})\cdot p(t)
\end{equation}

where $\pmb{u}_d$ denotes the unit color vector of the skin-tissue; $d_0$ denotes the stationary reflection strength; $\pmb{v}_{abs}(t)$ and $\pmb{v}_{sub}(t)$ denote the relative pulsatile strengths caused by both changes in hemoglobin and melanin absorption and changes in subsurface scattering respectively, as the blood volume changes; $p(t)$ denotes the BVP.

\begin{equation} \label{eq:3}
	\pmb{v}_s(t) = \pmb{u}_s \cdot (s_0+\Phi(m(t),p(t))) 
\end{equation}

where $\pmb{u}_s$ is the unit color vector of the light source spectrum; $s_0$ and $\Phi(m(t),p(t))$ denote the stationary and varying parts of specular reflections; $m(t)$ denotes all the non-physiological variations such as flickering of the light source, head rotation, facial expressions and actions (e.g.,~blinking, smiling).

\begin{equation} \label{eq:4}
	I(t) = I_0 \cdot (1+\Psi(m(t),p(t))) 
\end{equation}

where $I_0$ is the stationary part of the luminance intensity, and $I_0\cdot\Psi(m(t),p(t))$ is the intensity variation observed by the camera.

The interaction between physiological and non-physiological motions, $\Phi(\cdot)$ and $\Psi(\cdot)$, are usually complex non-linear functions.
The stationary components from the specular and diffuse reflections can be combined into a single component representing the stationary skin reflection:
\begin{equation} \label{eq:5}
	\pmb{u}_c \cdot c_0 = \pmb{u}_s \cdot s_0 + \pmb{u}_d \cdot d_0
\end{equation}
where $\pmb{u}_c$ denotes the unit color vector of the skin reflection and $c_0$ denotes the reflection strength. Substituting (\ref{eq:2}), (\ref{eq:3}), (\ref{eq:4}) and (\ref{eq:5}) into (\ref{eq:1}), produces:
\begin{multline} \label{eq:6}
	\pmb{C}_k(t)=I_0\cdot (1+\Psi(m(t),p(t))) \cdot \\
	(\pmb{u}_c \cdot c_0+\pmb{u}_s \cdot \Phi(m(t),p(t))+(\pmb{u}_{abs} + \pmb{u}_{sub}) \cdot p(t))+\pmb{v}_n(t)
\end{multline}

As the time-varying components are orders of magnitude smaller than the stationary components in (\ref{eq:6}), we can approximate $\pmb{C}_k(t)$ as:

\begin{multline} \label{eq:7}
	\pmb{C}_k(t)\approx \pmb{u}_c \cdot I_0 \cdot c_0+\pmb{u}_c \cdot I_0 \cdot c_0 \cdot \Psi(m(t),p(t)) + \\
	\pmb{u}_s \cdot I_0 \cdot \Phi(m(t),p(t))+(\pmb{u}_{abs} + \pmb{u}_{sub}) \cdot I_0 \cdot p(t)+\pmb{v}_n(t)
\end{multline}

For synthesizing data for physiological measurement methods, we want to create skin with RGB changes that vary with $p(t)$.
Using a principled bidirectional scattering distribution function (BSDF) shader, we are able to capture both of the components of $\pmb{u}_p$, $\pmb{u}_{abs}$ and $\pmb{u}_{sub}$, using the subsurface color and subsurface radius parameters. 
The specular reflections are controlled by the specular parameter. Thus, for a given pulse signal, $p(t)$, we can synthesize the skin's appearance over time. Furthermore, we can synthesize these changes in a wide variety of other variations, which for the purposes of vital sign measurement will represent noise sources. 

For any of the video-based physiological measurement methods, the task is to extract $p(t)$ from $\pmb{C}_k(t)$. The motivation for using a machine learning model to capture the relationship between $\pmb{C}_k(t)$ and $p(t)$ in (\ref{eq:7}) is that a neural model can capture a more complex relationships than hand-crafted demixing or source separation algorithms (e.g.,~ICA, PCA) that have ignored $p(t)$ inside $\Phi(\cdot)$ and $\Psi(\cdot)$, and assumed a linear relationship between $\pmb{C}_k(t)$ and $p(t)$.

\section{Avatar Synthesis}

We use high-fidelity facial avatars and a physiologically-based animation model 
for simulating videos of faces with a realistic blood flow (pulse) signal. These videos are then used to train a neural model for recovering the BVP from video sequences. The resulting model is tested on real video benchmark datasets. This process is shown in Fig.~\ref{fig:teaser}. 

\subsection{Synthetic Dataset}
We rendered nine video sequences for each of our 50 different facial identities, resulting in 450 video sequences in total. Each sequence was 10 seconds long, with a frame-rate of 30Hz.
The nine clips feature rotational head motions, facial expressions, and different backgrounds as described above. Each frame took approximately 20 seconds to render with Blender Cycles\footnote{https://docs.blender.org/manual/en/latest/render/cycles/index.html} on an Nvidia GTX 1080Ti GPU. 
These videos were used to train a convolutional attention network described in Section~\ref{sec:can}. The trained network was then tested on three benchmark video datasets of non-synthetic (a.k.a real) videos described in Section~\ref{sec:benchmarks}.

\subsection{Physiological Recordings}

To synthesize the appearance of the avatars, we use photoplethysmographic waveforms recordings from PhysioNet~\cite{goldberger2000physiobank}. Specifically, we use the BIDMC PPG and Respiration Dataset~\cite{pimentel2016toward} which include 53 8-minute contact PPG recordings sampled at 125Hz from different individuals.  These recordings were taken from the larger MIMIC-II dataset~\cite{saeed2011multiparameter}. We sample PPG recordings from different subjects for each of the 50 avatars that we synthesize. As we only synthesize/render short sequences (nine 10-second sequences described below) for each avatar we only use the first 90 seconds (9$\times$10 seconds) of each recording.

\subsection{Synthesizing Videos with Pulse Signals}

A key part of our work is a realistic model of facial blood flow. We simulate blood flow by adjusting properties of the physically-based shading material we use for the face\footnote{https://www.blender.org/}. 
The albedo component of the material is a texture map transferred from a high-quality 3D face scan.
The facial hair has been removed from these textures by an artist so that the skin properties can be easily manipulated (3D hair can be added later in the process). Specular effects are controlled with an artist-created roughness map, to make some parts of the face (e.g. the lips) shinier than others. An example of our material setup can be seen in Fig.~\ref{fig:approach}.

\textbf{Subsurface Skin Color:} As blood flows through the skin, the composition of the skin changes and causes variations in subsurface color. We manipulate skin tone changes using the subsurface color parameters. The weights for this are derived from the absorption spectrum of hemoglobin and typical frequency bands from an exemplar digital camera\footnote{\url{https://www.bnl.gov/atf/docs/scout-g_users_manual.pdf}} (Red: 550-700 nm, Green: 400-650 nm, Blue: 350-550 nm). In this work we globally vary these across all skin pixels on the albedo map (but not non-skin pixels).

\textbf{Subsurface Scattering:} We manipulate the subsurface radius for the channels to capture the changes in subsurface scattering as the blood volume varies. The subsurface scattering is spatially weighted using an artist-created subsurface scattering radius texture (see Fig.~\ref{fig:approach}) which captures variations in the thickness of the skin across the face. We vary the BSDF subsurface radii for the RGB channels using the same weighting prior as above.
Empirically we find these parameters work for synthesizing data for training camera-based vital sign measurement. We found that varying the subsurface scattering alone, without changes in subsurface color, were too subtle and could not recreate the effects the BVP on reflected light observed in real videos. 

\subsection{Systematic Variations}
To obtain machine learning systems that are robust to certain forms of variation encountered in the real world, we introduced the following types of variation into our dataset: 

\textbf{Facial Appearance.} We synthesized faces with 50 different appearances (examples can be seen in Fig.~\ref{fig:approach}). For each face, we set up the skin material with an albedo texture picked at random from our collection of 159 textures.
In order to model wrinkle-scale geometry, we also apply a matching high-resolution displacement map that was transferred from the scan data.  Skin type is particularly important in imaging PPG measurement. The approximate Fitzpartick skin type~\cite{fitzpatrick1988validity} distribution for the 50 faces was: Type I - 9, II - 15, III - 12, IV - 4, V - 5, VI - 5. While this distribution is still not uniform, it represents a much more balanced distribution than in existing imaging PPG datasets.
Just under half (21) of the avatars were synthesized with some form of facial hair (beard and/or moustache) to further increase the variety in appearance.

\textbf{Head Motion.} Since motion is one of the greatest sources of noise in imaging PPG measurement, we simulate a set of rigid head motions to augment training examples that capture these conditions. In particular, we smoothly rotate the head about the vertical axis at angular velocities of 0, 10, 20, and 30 degrees/second similar to prior work~\cite{estepp2014recovering}. Six of the nine videos synthesized for each avatar features motion, two at each angular velocity.

\textbf{Facial Expression.}
Similar to head motions, facial expressions movements are also a frequent source of noise in PPG measurement. We synthesized videos with smiling, blinking, and mouth opening (similar to speaking), which are some of the most common facial expressions exhibited in everyday life.
We apply smiles and blinks to the face using our collection of artist-created blend shapes, and we open the mouth by rotating the jaw bone with linear blend skinning. Four of the nine videos synthesized had smiling, mouth opening, and blinking motions.

\textbf{Environment.}
We render faces in different image-based environments to create a realistic variety in both background appearance and illumination on the face~\cite{debevec2006image}. For each sequence, we pick one high dynamic range spherical environment map from our collection~\cite{zaal2018hdri} (see Fig.~\ref{fig:approach} for examples).
In this work we synthesized static background scenes only, but future work may benefit from considering backgrounds with motion, or even facial occlusions that more closely resemble challenging real-life conditions.

\section{Experiments}

\subsection{Benchmark Datasets}
\label{sec:benchmarks}

\textbf{AFRL}~\cite{estepp2014recovering}: Videos were recorded at 658x492 pixel resolution and 120 frames per second (fps). Twenty-five participants (17 males) were recruited to participate in the study. Fingertip PPG was recorded as ground truth signals using a research-grade biopotential acquisition unit. Each participant was recorded six times for 5-minutes each with increasing head motion in each experiment and this process was repeated twice in front of two background screens. 

\textbf{MMSE-HR}~\cite{zhang2016multimodal}: 
102 videos of 40 participants were recorded at 25 fps capturing 1040x1392 resolution images during spontaneous emotion elicitation experiments. The gold standard contact signal was measured via a Biopac2 MP150 system\footnote{\url{https://www.biopac.com/}} which provided pulse rate at 1000 fps and was updated after each heartbeat. These videos feature smaller but more spontaneous motions than those in the AFRL dataset.

\textbf{UBFC-RPPG}~\cite{bobbia2019unsupervised}: 
42 videos of 42 participants were recorded at 640x480 resolution and 30 fps in uncompressed 8-bit RGB format. A fingertip oximeter was used to obtain the gold standard PPG.

\subsection{Physiological Measurement Network}
\label{sec:can}

To evaluate the impact of synthetic data on the quality of recovered pulse signals from video, we used an existing end-to-end learning model, Convolutional Attention Network (CAN)~\cite{chen2018deepphys}, which uses motion and appearance representations learned jointly through an attention mechanism. The approach consists of a two-branch convolutional neural network to represent motion and appearance.

The motion representation branch allows the network to differentiate between intensity variations caused by noise, e.g.,~from motion from subtle characteristic intensity variations induced by blood flow. The input to the motion representation branch is calculated as the difference of two consecutive video frames. To reduce the noise from changes in ambient illumination and the distance of the face to the illumination source, the frame difference is first normalized based on the skin reflection model~\cite{wang2017algorithmic}. The normalization is applied to each video sequence by subtracting the pixel mean and dividing by the standard deviation. We perform normalization on real and synthetic frames.  
 
The appearance representation captures the regions in the image that contribute strong iPPG signals. Via the attention mechanism, the appearance representation guides the motion representation and helps differentiate the PPG signal from the other sources of noise. The input frames are similarly normalized by subtracting the mean and dividing by the standard deviation. Again the same procedure is used for the real and synthetic frames.

\subsection{Training and Testing}

In all our experiments we use a person independent training regime and create training, validation and test partitions.

For experiments on the AFRL dataset, we perform a five-fold evaluation in which the 25 participants in the AFRL dataset~\cite{estepp2014recovering} were randomly divided into five folds, with 15 participants in the training set, five in the validation set, and five in the test set. The learning models were then trained to evaluate how our models can be generalized to new participants. The validation set was used to select the epoch for which the model would be used for testing. During training and model selection the mean squared error (MSE) between the predicted and gold-standard pulse waveforms was used as the loss/performance metric. 

The evaluation metrics for AFRL performance shown in the results tables are all averaged over the five folds. Prior work has shown that participant-independent training is a more challenging task than participant-dependent training~\cite{chen2018deepphys} and it is a more realistic scenario for real-world applications.
For experiments on the MMSE-HR and UBFC datasets, we use the model that performed best on the AFRL dataset and test it without fine-tuning (i.e., dataset independent evaluation). We compare our proposed approach to three other methods~\cite{poh2010non,CHROMdeHaan,wang2017algorithmic} for recovering the BVP. These methods are unsupervised and therefore results are reported across all participants without the need for cross-validation on either dataset.

For the convolutional neural network architecture motion representation model, we used nine layers with 128 hidden units, average pooling and tanh as the activation functions. The last layer of the motion model had linear activation units and the MSE loss. For the appearance model, we used the same architecture as the motion model but without the last three layers, consistent with~\cite{chen2018deepphys}. Finally, a 6th-order Butterworth filter was applied to all model outputs (cut-off frequencies of 0.7 and 2.5~Hz) before computing the frequency spectra and heart rate.
The baseline methods were implemented using the public MATLAB toolbox~\cite{mcduff2019iphys}.

\begin{table*}[hbt]
	\caption{Benchmark performance of pulse measurement on the AFRL~\cite{estepp2014recovering} and MMSE-HR~\cite{zhang2016multimodal} datasets.}
	\label{tab:AFRL_MMSE}
	\centering
	\small
	\setlength\tabcolsep{3pt} 
	\begin{tabular}{r|cccc|cccc|cccc}
	\toprule
		& \multicolumn{4}{c}{\textbf{AFRL (All Tasks)~\cite{estepp2014recovering} }} &  \multicolumn{4}{c}{\textbf{MMSE-HR~\cite{zhang2016multimodal}}} & \multicolumn{4}{c}{\textbf{UBFC~\cite{bobbia2019unsupervised}}} \\
        \textbf{Method} & MAE & RMSE  & $\rho$ & SNR & MAE & RMSE  & $\rho$ & SNR & MAE & RMSE  & $\rho$ & SNR \\ \hline \hline
        CAN (w/ Real+Synth) & \textbf{2.42} & \textbf{4.37} & 0.88 & \textbf{6.57} & \textbf{2.26} & \textbf{3.70} & \textbf{0.97} & \textbf{4.85}  &  5.55 & 12.9 & 0.66 & \textbf{0.70} \\ 
        CAN (w/ Synth) & 9.23  & 13.4 & 0.36 & -7.17 & 4.98 & 11.8 & 0.70 & -1.98 & \textbf{5.16} & \textbf{10.1} & \textbf{0.80} & -2.83  \\
        CAN (w/ Real)~\cite{chen2018deepphys} & 2.43 & 4.39 & 0.87 & 6.21 & 4.43 & 9.98 & 0.80 & -0.66 & 5.58 & 11.8 & 0.71 & -0.90  \\
        POS~\cite{wang2017algorithmic} & 2.48 & 5.07 & \textbf{0.89} & 2.32 & 3.90 & 9.61 & 0.78 & 2.33 & 8.24 & 19.9 & 0.57 & -1.19  \\
        CHROM~\cite{CHROMdeHaan} & 6.42 & 12.4  & 0.60  & -4.83 & 3.74 & 8.11 & 0.82 & 1.90 & 7.46 & 15.5 & 0.72 & -1.10 \\
        ICA~\cite{poh2010non} & 4.36 & 7.84 & 0.77 & 3.64 & 5.44 & 12.00 & 0.66 & 3.03 & 14.3 & 26.1 & 0.28 & -2.67 \\ \bottomrule
   \end{tabular}
   \\
   \tiny
   MAE = Mean Absolute Error in HR estimation, RMSE = Root Mean Squared Error in HR estimation, SNR = BVP Signal-to-Noise Ratio, $\rho$ = Pearson Correlation in HR estimation.
\end{table*}

\begin{figure*}[h]
  \centering
  \includegraphics[width=1\linewidth]{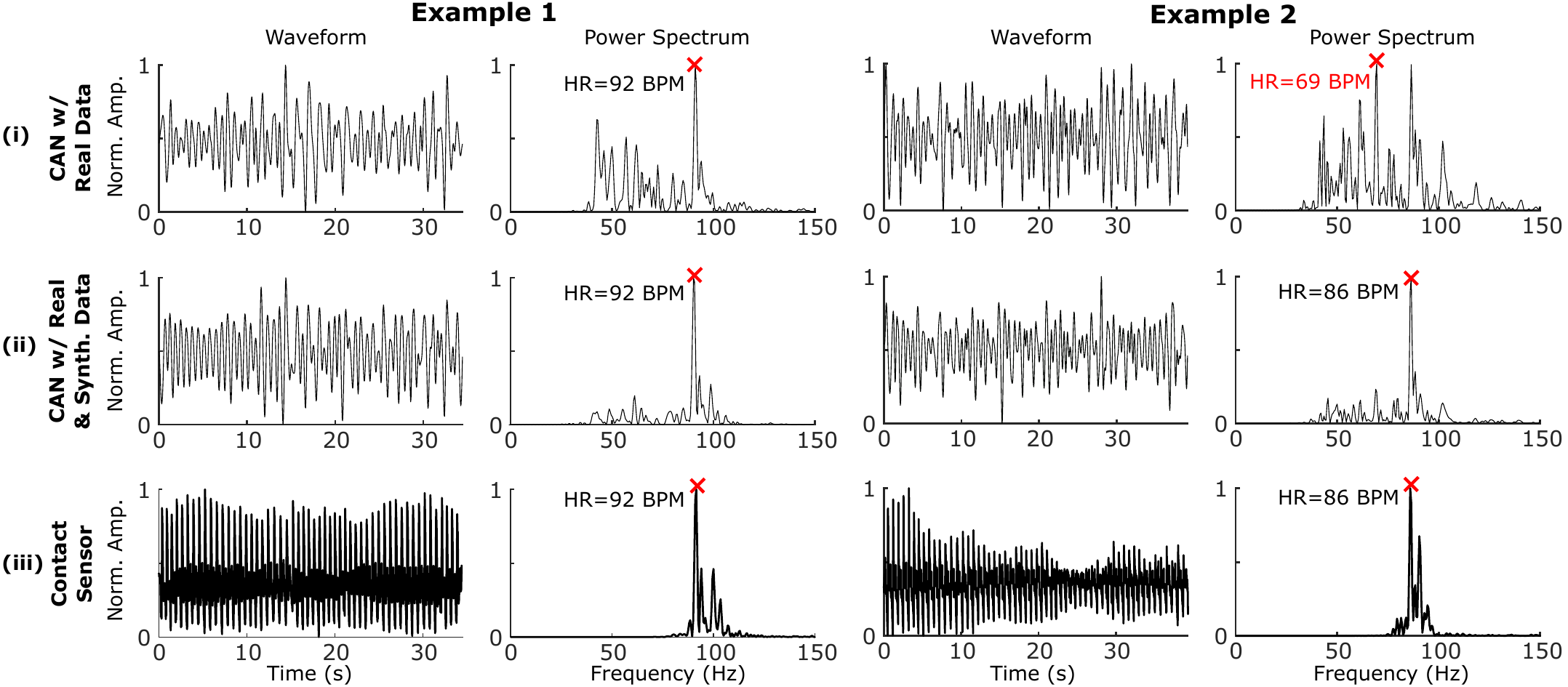}
  \caption{Examples of the blood volume pulse recovered using a camera and the trained neural network (thin black lines), with only (i) real and (ii) synthetic and real data, in comparison with (iii) a finger contact sensor (thick black lines). PPG waveforms normalized from 0 to 1 and pulse power spectra normalized from 0 to 1 are shown.  Notice how the pulse spectra are much cleaner and more closely resemble the gold-standard when using the model trained on real and synthetic data, compared to real data alone. These are qualitative examples of why we see an improvement in BVP SNR as shown in Tables~\ref{tab:AFRL_MMSE}.}
  \label{fig:waveform_examples}
\end{figure*}

\section{Results}

\textbf{Training with Synthetic Data}.
Our first experiments are to validate the effect of using synthetic data to train the vital signs measurement algorithm. Table~\ref{tab:AFRL_MMSE} shows results of models trained on non-synthetic (real) data, synthetic data, and a combination of real and synthetic data. 
Results are shown for the AFRL dataset for which we perform the five-fold participant-independent cross-validation. For the MMSE results we report performance of the model trained on the AFRL data and thus this is both participant independent (because no people feature in both dataset) and can be viewed as an example of cross-dataset transfer learning. 
The models trained on real and synthetic data outperform the models trained only on real data for both datasets. This is true for the BVP SNR reflecting that the underlying pulse signal is cleaner and for HR MAE and RMSE reflecting that the HR estimates are more accurate. On the AFRL dataset the results are not very different because training with real data from the same dataset already performs very well (and the HR correlation is marginally higher). This is because synthetic data does not provide a great benefit if the distribution of the test data is very similar to that of the training data. In the AFRL dataset the participants all have similar skin types (tones) and the lighting is very constant across all videos. Thus, if examples of all tasks from this dataset are included in the training set, even if they feature different participants, the margin for improvement is small. 

However, when we perform a cross-dataset using a combination of synthetic and real data set provides a much more considerable improvement. The MAE in HR estimates on the MMSE dataset is 2.26 (compared to 3.74 for the next best approach), a 40\% reduction in error. In this case, the distribution of the testing data is quite different from the data in the training set and, consequently, the benefit of using synthetics becomes apparent. The synthetic data help improve the generalization of the model when there is a larger domain gap between the training and testing data. To provide a qualitative example, Fig.~\ref{fig:waveform_examples} shows an example of the recovered pulse waveforms and corresponding power spectra for two videos in the MMSE-HR dataset. Notice how the pulse spectra are much cleaner and more closely resemble the gold-standard when using the model trained on real and synthetic data. Interestingly, the performance on the UBFC dataset is strongest when training with only synthetic data, we hypothesize this is because the domain gap between the real training data and the UBFC test data is larger. Future work will investigate how to characterize the difference between dataset distributions in this domain.

\textbf{Cross-Task Performance.}
Body motions are one of the most common and problematic sources of noise in non-contact vital signs measurement.
In the previous analyses on the AFRL dataset we included examples of every task in the training, validation and test sets. However, when we train and validate only on videos with static subjects and then test on videos with head motions the improvements gained from using synthetic data are much more dramatic (shown in Table~\ref{tab:static_motion}). The avatar data includes heads with motions, the result highlights that synthetic data can help bridge the gap between heads with motion. For example, if no real video data with gold-standard measurements were available with motions similar to those in the test scenario we can synthesize data to bridge the gap.

\textbf{Comparison with Benchmarks.}
Next let us compare performance on both datasets against the other baseline methods. Table~\ref{tab:AFRL_MMSE} shows the results of the CAN alongside ICA~\cite{poh2010non}, CHROM~\cite{de2013robust} and POS~\cite{wang2017algorithmic}. On both datasets the neural network trained on real and synthetic data outperforms all the other methods. The POS method performs well on the AFRL dataset with similar results in HR estimation, but a lower BVP SNR. On the MMSE dataset the CAN outperforms the other methods by a considerable margin (MAE = 2.26 BPM vs. 3.74 BPM from the next best method). Unsupervised methods have previously been used more frequently than supervised algorithms for imaging-based measurement of vital signs because of concerns about the generalizability of ``trained'' models. However, our results suggest that with sufficient diversity in the training set that supervised methods could have an advantage. 
\textbf{Robustness to Skin Tone.}
Skin type influences the signal-to-noise ratio (SNR) of the recovered BVP in many camera-based vital sign measurement algorithms~\cite{addison2018video,shao2016simultaneous, wang2017algorithmic}. A larger melanin concentration in people with darker skin absorbs more light, making the intensity of light returning to the camera lower and thus the iPPG signal weaker. To exacerbate this problem subjects with darker skin types are often underrepresented in computer vision datasets, including those used for camera-based physiological measurement. Synthetic data can be used to identify biases in CV systems and help address them~\cite{mcduff2019identifying}.

\begin{figure}[h]
  \centering
  \includegraphics[width=1\linewidth]{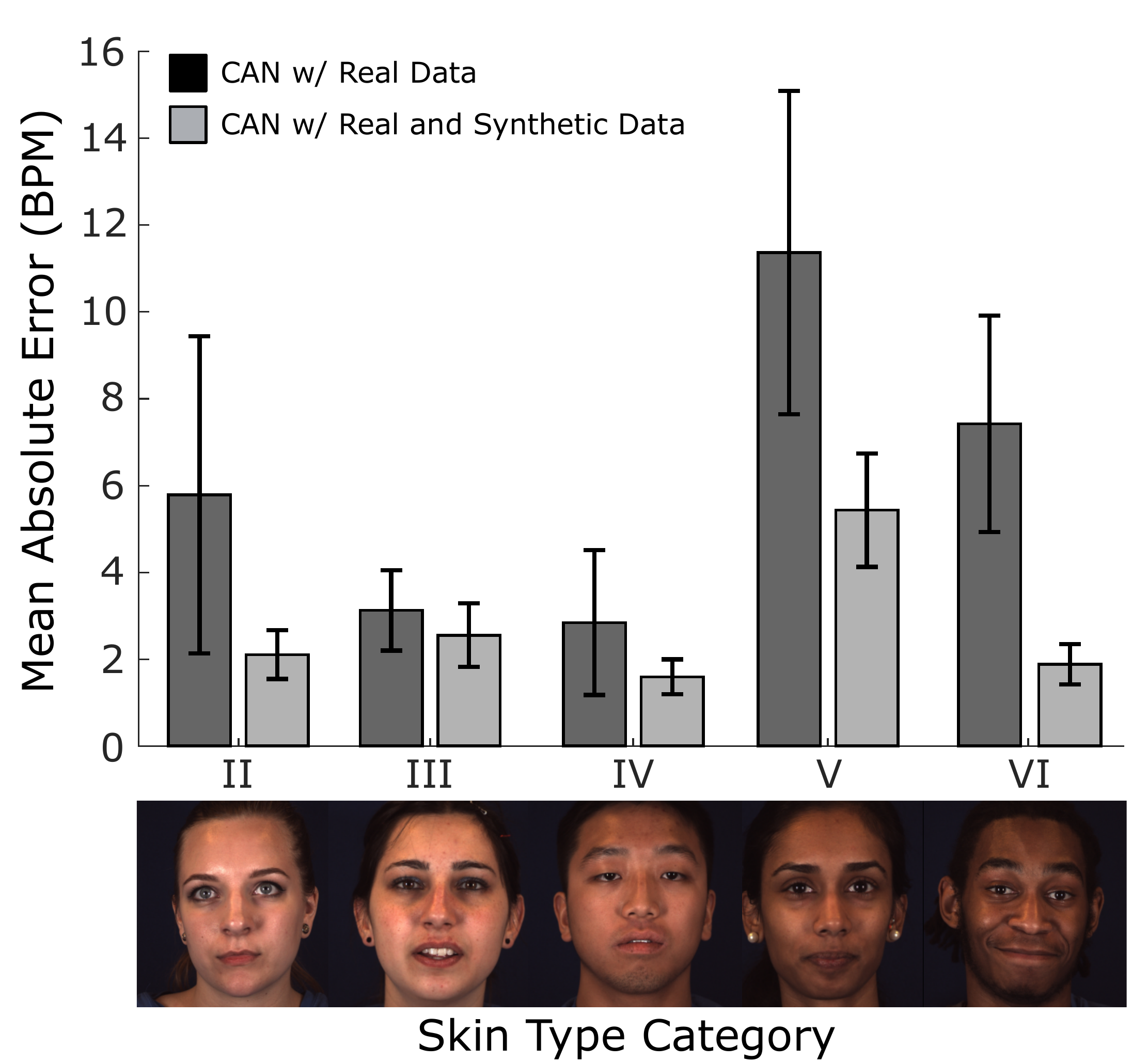}
  \caption{Heart rate mean absolute error (BPM) by skin tone on the MMSE-HR dataset~\cite{zhang2016multimodal}. Training with synthetic data reduces the errors for the lightest (II) and darkest (VI) skin types the most, those that are often underrepresented in real video training datasets. No. of participants: II=8, III=11, IV=18, V=2, VI=2.}
  \label{fig:skintone}
\end{figure}

Table~\ref{fig:skintone} shows the performance when testing the model on subjects with different skin types based on the Fitzpatrick skin type scale~\cite{fitzpatrick1988validity} from the MMSE-HR dataset~\cite{zhang2016multimodal}. The synthetic data provides a substantial improvement in HR MAE, especially for the lightest (II) and darkest (VI) skin types.  These are the skin types that are typically underrepresented in real video datasets used for non-contact vital sign measurement algorithms, including the AFRL dataset. Not only are the overall heart rate estimation errors lower for all skin types, the standard deviation in average heart rate MAE across skin types is approximately halved when training with real and synthetic data compared to when training with only real data. To summarize, our results show that by using synthetic data we can also improve performance on subjects whose appearance type (in this case skin type) was underrepresented in the ``real" portion of the training data set.

\begin{table}[hbt]
	\caption{Task-Independent Performance: Pulse measurement on videos without head motion (Tasks 1 \& 2) and with head motion (Tasks 3, 4, 5 \& 6) from the AFRL~\cite{estepp2014recovering} dataset when training on videos without motion (Tasks 1 \& 2).}
	\label{tab:static_motion}
	\centering
	\small
	\setlength\tabcolsep{3pt} 
	\begin{tabular}{r|cccc}
	\toprule
		& \multicolumn{4}{c}{\textbf{AFRL Motion Tasks (3-6)}} \\
        \textbf{Method} & MAE & RMSE  & SNR & $\rho$ \\ \hline \hline
        CAN (w/ Real (Static) + Synthetic) & \textbf{6.52} & \textbf{9.82} & \textbf{-0.30} & \textbf{0.63}   \\
        CAN (w/ Real (Static)) & 8.21 & 11.8 & -1.68 & 0.50   \\ 
        CAN (w/ Synthetic) & 14.1 & 18.9 & -10.2 & 0.16 \\ \bottomrule
   \end{tabular}
   \\
   \tiny
   MAE = Mean Absolute Error in HR estimation, RMSE = Root Mean Squared Error in HR estimation, SNR = BVP Signal-to-Noise Ratio, $\rho$ = Pearson Correlation in HR estimation, WMAE = Waveform MAE.
\end{table}

\begin{table}[hbt]
	\caption{Mean absolute error in heart rate estimation by participant Fitzpatrick skin tone category.  The standard deviation in average errors across skin tones is shown in the final column.}
	\label{tab:skintone}
	\centering
	\small
	\setlength\tabcolsep{3pt} 
	\begin{tabular}{r|ccccc|c}
	\toprule
		& \multicolumn{5}{c}{\textbf{Fitz. Skin Tone}} \\
        \textbf{Method} & II  & III & IV & V & VI & $\sigma$ \\ \hline \hline
        CAN (w/ Real + Synthetic) & 2.10 & 2.55 & 1.59 & 5.43 & 1.88 & 1.56  \\
        CAN (w/ Real) & 5.79 & 3.12 & 2.84 & 11.4 & 7.42 & 3.50   \\ 
        CAN (w/ Synthetic) & 2.60 & 2.83 & 3.47 & 13.9 & 31.8 & 12.6 \\ \bottomrule
   \end{tabular}
   \\
\end{table}

\section{Discussion}

Collecting datasets for training non-contact vital signal measurement algorithms has several challenges. We have presented an approach for synthesizing avatars that helps alleviate the need for real videos. Our results show that training with synthetic data can successfully improve the performance of non-contact vital sign measurement. Specifically, including synthesized and real video data in the training set can lead to an improvement of the pulse SNR ratio as well as lowering heart rate measurements errors compared to training with just real video data alone. We achieve state-of-the-art performance on three large video datasets. In particular, the recovered BVP signal quality was much improved across both datasets (see Table~\ref{tab:AFRL_MMSE}). 

When training and testing on the same datasets (in a participant-independent manner), the improvements were modest. This suggests that when the testing data has a similar distribution (similar motions, lighting, skin types) there is not much benefit to be gained from synthetic data. Synthetic data is particularly effective at reducing errors in cross-domain learning, improving cross-task, cross-dataset and cross-appearance generalization. Synthesizing data allows us to create many combinations of facial appearances (skin tones, hair styles, facial hair), expressions, speech, head motions (rotational and translational), ambient lighting conditions and backgrounds. Finally, we show that synthetic data can substantially improve (and reduce the variance in) performance of non-contact vital sign measurement for skin tones underrepresented in training data. 

While synthetics are flexible and scalable once you have created a pipeline, the initial overhead for this infrastructure is expensive and labor-intensive to create.  Our synthetics pipeline involved a multi-year effort to create. Furthermore, while we demonstrate that our synthetics pipeline can offer a tangible benefit, we did not push the limits of the improvements that synthetic data can provide. It is possible that greater improvements could have been obtained if we had synthesized more face videos. However, the videos were synthesized on a frame-by-frame basis taking approximately an hour to synthesize a single 10s video.

While our approach could be used for creating more motion robust iPPG algorithms for non-contact measurement in fitness centers or telehealth systems. Many of the applications of non-contact vital signal measurement do not necessarily involve analysis of adult faces. Modeling infants for training models to be deployed in a NICU would be a great extension of this work.

\section{Conclusion}
This work proposes the use of synthetic avatars to synthesize novel samples of facial blood volume changes that can improve the robustness of non-contact physiological sensing methods. We are looking forward to a future when similar methodology can be used to not only improve the generalization performance under challenging real-life scenarios but also minimize potential performance differences across underrepresented groups or people.

\balance{}

{\small
\bibliographystyle{ieee_fullname}
\bibliography{egbib}
}

\end{document}